\documentclass[letterpaper, 10 pt, conference]{ieeeconf}
\let\labelindent\relax

\usepackage{booktabs}
\usepackage{amsmath,amssymb,bm}
\usepackage{graphicx}
\usepackage{multirow}
\usepackage[table]{xcolor}
\usepackage{enumitem}

\makeatletter
\let\NAT@parse\undefined
\makeatother

\usepackage[colorlinks,
            linkcolor=red,
            anchorcolor=blue,
            urlcolor=magenta,
            pdfstartview=FitH, 
            hypertexnames=false,
            citecolor=blue]{hyperref}
\usepackage[noadjust]{cite}

\IEEEoverridecommandlockouts
\title{\LARGE \bf
PanoGS-SLAM: Panoramic 3D Gaussian Splatting SLAM
}

\author{Yongqi Mao$^{1}$, Hao Shi$^{3, 1}$, Yufan Zhang$^{2}$, Zhonghua Yi$^{1}$, Xiangfei Guo$^{1}$, Kaiwei Wang$^{1*}$\\$^{1}$Zhejiang University, $^{2}$National University of Defense Technology, 
$^{3}$Ant Group
\thanks{$^{*}$Corresponding author: Kaiwei Wang (E-mail: wangkaiwei@zju.edu.cn)}
\thanks{This research was supported by Zhejiang Provincial Natural Science Foundation of China under Grant No. LZ24F050003}}

\begin{document}

\maketitle
\thispagestyle{empty}
\pagestyle{empty}

\begin{abstract}
Real-time dense SLAM is a core capability for robotics applications that require robust localization and high-quality mapping in dynamic or fast-changing environments. Recent 3D Gaussian Splatting (3DGS)-based SLAM methods have shown promising performance, but most are designed for narrow-FoV pinhole cameras, where limited angular coverage weakens pose observability and often leads to unstable photometric optimization under rapid motion and large viewpoint changes. 
We present \textbf{PanoGS-SLAM}, the first panoramic dense SLAM system built on 3D Gaussian Splatting. Our method performs differentiable rendering and pose optimization directly in the spherical domain, enabling omnidirectional photometric constraints for more stable tracking. To improve geometric consistency and robustness, we introduce (1) a sphere-consistent photometric loss that compensates for the area distortion of equirectangular projection, and (2) a depth-guided Gaussian initialization strategy that stabilizes incremental mapping in newly observed regions.
Extensive experiments on both real and synthetic panoramic benchmarks (PALVIO and SynPano) show that PanoGS-SLAM consistently outperforms geometric and GS-based baselines in tracking accuracy and rendering quality, while achieving fast front-end convergence and real-time performance. 
In addition, controlled field-of-view experiments reveal a clear monotonic improvement in optimization conditioning and convergence stability as angular coverage increases, highlighting the fundamental role of sensing geometry in shaping the optimization landscape of differentiable Gaussian-based SLAM. The source code will be made publicly available.
\end{abstract}

\section{Introduction}
Real-time dense SLAM is a fundamental capability for robotic systems that require reliable localization, scene understanding, and high-fidelity mapping, such as autonomous navigation, 3D reconstruction, and immersive perception. 
Classical dense SLAM pipelines typically rely on explicit scene representations (e.g., depth maps\cite{Newcombe2011DTAM}, voxels\cite{BundleFusion}\cite{KinectFusion}, or surfels\cite{Point-BasedFusion}\cite{ElasticFusion}), which often face an unfavorable trade-off between scalability, efficiency, and rendering fidelity. 
Recent advances in differentiable scene representations, especially 3D Gaussian Splatting (3DGS)\cite{kerbl3Dgaussians}, provide a compelling alternative by combining explicit geometric primitives with efficient differentiable rendering.
Building on 3DGS, several recent methods have extended Gaussian representations to dense visual SLAM\cite{yan2023gs}\cite{Matsuki:Murai:etal:CVPR2024}\cite{keetha2024splatam}\cite{hhuang2024photoslam}\cite{gs-icp}\cite{vings}. 
By jointly optimizing camera poses and Gaussian primitives through photometric rendering losses, these systems achieve a favorable balance between real-time performance and reconstruction quality. 
This line of work has established Gaussian-based differentiable SLAM as a promising paradigm for high-quality online mapping.

\begin{figure}[!t]
  \centering
  \includegraphics[width=\linewidth]{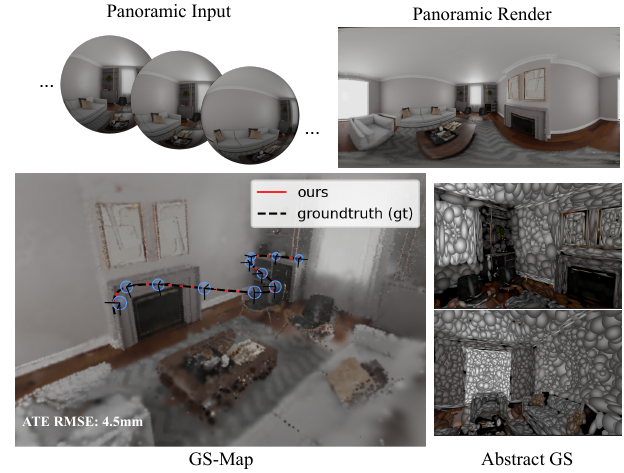}
  \caption{\textbf{PanoGS-SLAM} takes monocular panoramic input and performs differentiable rendering and pose optimization directly in the spherical domain. The large FoV provides strong constraints, enabling fast and stable pose convergence with millimeter-level trajectory accuracy.}
  \label{fig:teaser}
\end{figure}
However, existing GS-based SLAM systems are predominantly developed under the narrow-FoV pinhole camera assumption. 
This design choice introduces a fundamental limitation for differentiable photometric pose estimation: the optimization landscape is strongly influenced by the spatial and angular distribution of image gradients.
Under limited FoV, gradients are concentrated within a narrow viewing cone, leading to an anisotropic information distribution, weakened rotational observability, and stronger coupling between translational and rotational degrees of freedom. As a consequence, the underlying optimization problem becomes poorly conditioned, exhibiting a reduced convergence basin and increased sensitivity to rapid motion or abrupt viewpoint changes.
A second issue arises during incremental mapping. 
Under narrow-FoV sensing, newly observed regions are introduced frequently, but these regions initially lack sufficient Gaussian support and geometric constraints. 
This weakly constrained state can destabilize front-end pose optimization and slow down convergence, especially when tracking depends on a lagging global map. 
In practice, this behavior is evident in recent monocular GS-based systems such as MonoGS\cite{Matsuki:Murai:etal:CVPR2024}, where tracking quality can degrade significantly when the camera enters large unseen areas.

In this work, we revisit Gaussian-based dense SLAM from the perspective of sensing geometry. We hypothesize that wide-field perception fundamentally reshapes the optimization landscape of differentiable SLAM by redistributing photometric constraints across the full viewing sphere. To validate this hypothesis, we present \textbf{PanoGS-SLAM}, the first panoramic SLAM system built upon 3D Gaussian Splatting. Instead of decomposing panoramas into perspective crops, our framework performs differentiable rendering and joint optimization directly in the spherical domain, maintaining continuous gradients over the entire unit sphere. This design substantially improves numerical conditioning and enlarges the convergence basin of pose optimization.
To ensure geometric consistency under equirectangular projection, we introduce a sphere-consistent photometric objective that compensates for non-uniform pixel area distortion, preventing polar regions from dominating the loss function.
We further propose a depth-guided Gaussian initialization and insertion strategy to enhance geometric priors in newly observed regions and stabilize incremental mapping. 
Together, these components form a unified panoramic Gaussian SLAM framework for robust tracking and high-quality reconstruction from omnidirectional RGB input.

Extensive experiments on both synthetic and real panoramic benchmarks (SynPano\cite{synpano2026} and PALVIO\cite{wang2022lf}) demonstrate the effectiveness of our approach. 
PanoGS-SLAM consistently achieves the best tracking accuracy and superior rendering quality over both geometric panoramic SLAM baselines and recent GS-based dense SLAM methods. On several sequences, it outperforms existing baselines by up to an order of magnitude.
Moreover, our system converges much faster in front-end pose optimization, requiring only 15 iterations to achieve stable tracking—approximately one-seventh of the iterations required by MonoGS\cite{Matsuki:Murai:etal:CVPR2024}. This leads to substantially improved runtime efficiency.
Beyond benchmark comparisons, we conduct a controlled field-of-view study and show a clear monotonic trend: increasing angular coverage markedly improves convergence behavior and numerical stability in differentiable pose optimization. 
These results reveal that the field of view is not only a sensing choice but also a key factor governing the optimization properties of GS-based dense SLAM.

Our contributions are summarized as follows:
\begin{itemize}
    \item \textbf{Panoramic 3D Gaussian SLAM Framework.}
    We present the first 3DGS-based panoramic SLAM system that directly performs joint camera tracking and Gaussian map optimization in the spherical domain.

    \item \textbf{Sphere-Consistent Photometric Optimization.}
    A geometrically consistent panoramic loss compensating for equirectangular area distortion to improve gradient balance and tracking stability.
    
    \item \textbf{Depth-Guided Gaussian Initialization For Incremental Mapping.}
    A depth-guided Gaussian insertion strategy that enhances geometric consistency in newly observed regions.

    \item \textbf{Field-of-View Analysis For Differentiable SLAM.}
    Through controlled FoV experiments, we empirically show that wider angular coverage substantially improves conditioning, convergence basin, and robustness in Gaussian-based dense SLAM.
\end{itemize}

\section{RELATED WORK}
\subsection{Dense SLAM}
Traditional sparse SLAM systems achieve remarkable performance in camera localization and mapping, as demonstrated by ORB-SLAM\cite{mur2015orb}\cite{mur2017orb}\cite{ORBSLAM3_TRO} and VINS-Mono\cite{qin2017vins}. However, the resulting maps are typically sparse and mainly optimized for pose estimation. In contrast, dense SLAM aims to reconstruct richer scene representations, providing enhanced perception for augmented reality and robotic applications.
Dense SLAM systems fuse per-pixel observations to recover continuous surfaces, commonly using voxel-based representations\cite{BundleFusion}\cite{KinectFusion} or point-based models\cite{ElasticFusion}\cite{Point-BasedFusion}. Although effective, these explicit representations often incur high memory consumption and limited rendering quality.
NeRF\cite{mildenhall2021nerf} introduced neural implicit representations for modeling scene geometry and appearance. Building on this idea, iMAP\cite{iMAP} incorporated implicit neural maps into SLAM, while NICE-SLAM\cite{zhu2022nice} improved scalability through a hierarchical multi-resolution design. Subsequent works combined geometric representations with neural fields to enhance rendering fidelity\cite{yang2022vox}\cite{sandstrom2023point}, but these approaches generally require time-consuming optimization, limiting real-time applicability.
More recently, 3DGS\cite{kerbl3Dgaussians} proposed an explicit and differentiable representation that significantly reduces optimization time while maintaining high rendering quality. Methods such as SplaTAM\cite{keetha2024splatam}, GS-SLAM\cite{yan2023gs}, MonoGS\cite{hhuang2024photoslam}, and Photo-SLAM\cite{Matsuki:Murai:etal:CVPR2024} extended 3DGS to dense SLAM by jointly optimizing poses and Gaussian parameters, improving robustness and photometric consistency. GS-ICP\cite{gs-icp} efficiently registers Gaussian spheres with an ICP-like optimization, and VINGS-SLAM\cite{vings} integrates visual-inertial odometry with GS–based mapping. Nevertheless, these methods are largely based on the pinhole camera model, whose limited field of view may compromise tracking stability under rapid viewpoint changes.

\subsection{Omnidirectional SLAM and 3DGS}
Omnidirectional images provide richer geometric constraints due to their wide field of view, which generally improves tracking robustness under rapid motion and abrupt scene transitions. OpenVSLAM\cite{sumikura2019openvslam} extends ORB-SLAM2\cite{mur2017orb} to support panoramic cameras. Building upon ORB-SLAM3\cite{ORBSLAM3_TRO}, P2U-SLAM\cite{zhang2026p2u} further proposes a unified omnidirectional framework that explicitly models point and pose uncertainties. LF-VISLAM\cite{LF-VISLAM}, derived from VINS-Mono\cite{qin2017vins}, introduces a tightly coupled visual-inertial framework with loop closure capability that accommodates omnidirectional inputs. 360VO\cite{360VO} extends the classical Direct Sparse Visual Odometry framework to the omnidirectional camera setting.
360-GS\cite{bai2025360} extends 3DGS to omnidirectional scene modeling and demonstrates global reconstruction for panoramic images. 
Subsequent works such as ODGS\cite{lee2024odgs} and OmniGS\cite{OmniGS} introduce dedicated CUDA rasterizers that directly operate in spherical or equirectangular projection spaces, avoiding intermediate perspective decomposition and improving rendering efficiency.
However, these methods focus on offline reconstruction and do not address camera pose estimation or global consistency in an online SLAM setting. 
Consequently, the integration of omnidirectional 3DGS into a globally consistent SLAM framework persists as a significant, unresolved challenge. In light of this, we present a novel, globally consistent omnidirectional 3DGS-based SLAM system that concurrently optimizes both camera poses and Gaussian parameters within a cohesive, unified framework.

\begin{figure*}[thpb]
\centering
\includegraphics[width=\linewidth]{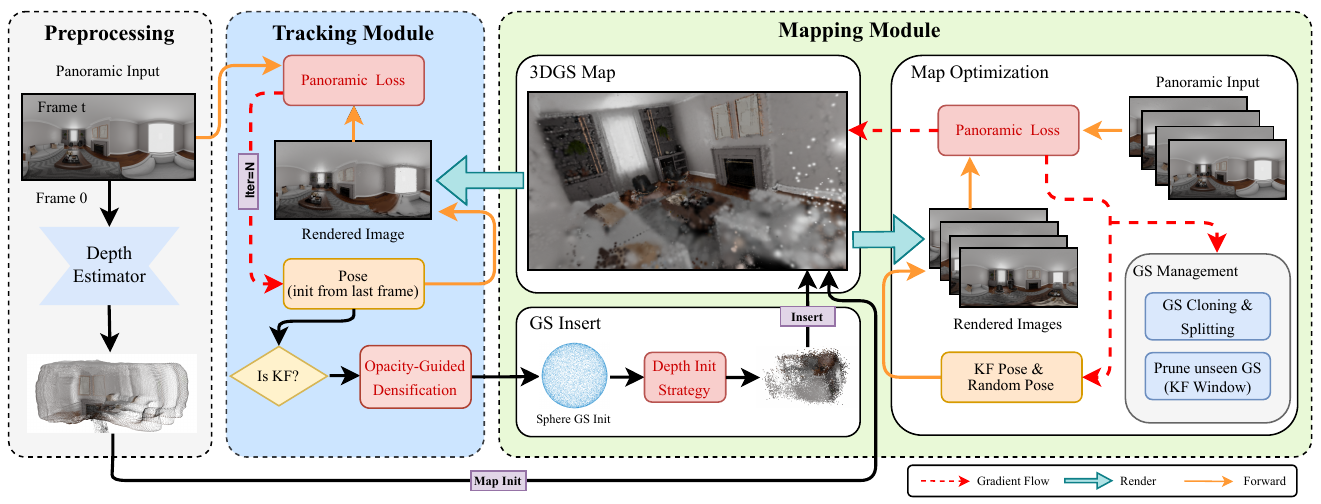}
\caption{\textbf{PanoGS-SLAM Pipeline.} The system consists of preprocessing, tracking, and mapping modules, operating on panoramic inputs with 3DGS as the sole representation. The initial map is constructed from the first frame using a pre-trained panoramic depth estimator\cite{BiFuse20}. 
In tracking, camera poses are estimated by minimizing the panoramic loss ${L}_{pano}$ on the unit sphere, initialized using the previous-frame pose. 
Keyframes trigger opacity-guided Gaussian insertion. 
In mapping, newly inserted Gaussians are generated via equal-area unit-sphere sampling and depth initialization. 
The back-end jointly optimizes keyframe poses, randomly sampled non-keyframes, and Gaussian parameters within the keyframe window, while GS management performs cloning, splitting, and pruning to maintain map quality.}
\label{fig:pipline}
\end{figure*}
\section{METHOD}
The overall architecture of PanoGS-SLAM is illustrated in Fig.~\ref{fig:pipline}. Our system takes monocular panoramic sequences as input and maintains a unified 3D Gaussian representation for both tracking and mapping. The pipeline consists of a preprocessing module for initial depth estimation, a tracking front-end for pose optimization, and a mapping back-end for incremental scene reconstruction. In the following, we describe the core components of our framework.

\subsection{Gaussian Splatting}
We adopt 3DGS as the scene representation, denoted as $\{G_i\}_{i=1}^N$. Each Gaussian is parameterized by its mean position $\boldsymbol{\mu}_i \in \mathbb{R}^3$ in the world coordinate frame, opacity $\alpha_i \in [0,1]$, and covariance matrix $\boldsymbol{\Sigma}_i \in \mathbb{R}^{3 \times 3}$:
\begin{equation}
G_i(\mathbf{x}) = \exp\!\left(-\tfrac{1}{2}\,(\mathbf{x}-\boldsymbol{\mu}_i)^\top \boldsymbol{\Sigma}_i^{-1} (\mathbf{x}-\boldsymbol{\mu}_i)\right).
\end{equation}
In addition, the appearance of each Gaussian is modeled using spherical harmonics (SH) to represent view-dependent color.
Before splatting, each 3D Gaussian is projected onto the 2D image plane, yielding a 2D covariance matrix:
\begin{equation}
\boldsymbol{\Sigma}_{\mathrm{2D}}
=
\mathbf{J}\mathbf{W}
\boldsymbol{\Sigma}
\mathbf{W}^{\top}\mathbf{J}^{\top}.
\label{cov2D}
\end{equation}
where $\mathbf{J}$ denotes the Jacobian of the projection function and $\mathbf{W}$ is the camera pose transformation matrix.
3DGS performs volumetric rendering without explicit geometric surfaces. 
The final color at each pixel is synthesized by compositing $N$ Gaussians along the viewing direction:
\begin{equation}
C = \sum_{i=1}^{N} c_i \alpha_i \prod_{j=1}^{i-1} (1-\alpha_j).
\end{equation}
where $c_i$ is the color of the $i$-th Gaussian.
Unlike ray-based volume rendering, 3DGS adopts a rasterization-based pipeline that traverses 2D Gaussians on the image plane for each pixel. This design fully exploits the sparsity of the 3D scene representation and enables highly efficient rendering.

\subsection{Camera Model}
Standard 3DGS is primarily designed for perspective projection. To adapt it for omnidirectional SLAM, we replace the pinhole model with a spherical projection that maps 3D Gaussians onto an equirectangular projection (ERP) plane, following the prior panoramic Gaussian Splatting method\cite{lee2024odgs}. Let $\boldsymbol{\mu} = (\mu_x, \mu_y, \mu_z)^\top$ denote the mean position of a Gaussian in the camera coordinate frame. Its longitude $\phi_\mu$ and latitude $\theta_\mu$ in spherical coordinates are defined as:
\begin{equation}
\phi_\mu=\arctan\!\left(\frac{\mu_x}{\mu_z}\right),
\end{equation}
\begin{equation}
\theta_\mu=\arctan\!\left(\frac{-\mu_y}{\sqrt{\mu_x^2+\mu_z^2}}\right).
\end{equation}
The spherical coordinates are then mapped to pixel coordinates $(u, v)$ via the cylindrical projection function $\pi_o(\cdot)$:
\begin{equation}
\pi_o\left(\boldsymbol{\mu}\right)=\left(\frac{W}{2\pi}\phi_\mu+\frac{W}{2},\ -\frac{H}{\pi}\theta_\mu+\frac{H}{2}\right)^\top.
\end{equation}
Here, $W$ and $H$ are the width and height of the panoramic image. To propagate covariance information during rasterization, we compute the Jacobian matrix $\mathbf{J}$ of the projection $\pi_o$ by differentiating it with respect to the 3D Gaussian position. This Jacobian captures the local geometric deformation induced by the mapping from Euclidean space to the spherical domain, ensuring that the 3D Gaussians are properly stretched near the poles of the panoramic image:
\begin{equation}
\begin{aligned}
a = \frac{W}{2\pi\lVert \boldsymbol{\mu} \rVert}, b = \frac{H}{\pi\lVert \boldsymbol{\mu} \rVert},
\end{aligned}
\end{equation}
\begin{equation}
\mathbf{J} =
\begin{pmatrix}
a\sec\theta_\mu\cos\phi_\mu & 0 & -a\sec\theta_\mu\sin\phi_\mu \\
b\sin\theta_\mu\sin\phi_\mu & b\cos\theta_\mu & b\sin\theta_\mu\cos\phi_\mu
\end{pmatrix}.
\end{equation}
The final 2D covariance is computed as $\boldsymbol{\Sigma}_{\mathrm{2D}}$ in \eqref{cov2D}, completing the transformation from 3D Gaussians to 2D Gaussians on the ERP plane. The resulting 2D Gaussians are then rasterized and blended to synthesize pixel colors, following the standard Gaussian splatting procedure.

\subsection{SLAM Front-end}
In the front-end, we optimize only the camera pose while keeping the Gaussian map fixed. 
Unlike narrow-FoV pinhole cameras, which often exhibit discontinuous pose gradients near image boundaries, the panoramic camera provides omnidirectional coverage. This ensures smooth gradients across the entire view, resulting in a more stable and efficient pose optimization.
To further enhance tracking stability, we design a sphere-aware Panoramic Loss tailored to panoramic images and incorporate a keyframe-based management strategy to maintain map consistency.

\textbf{Panoramic Loss.}
Equirectangular images introduce area distortion: pixels near the poles cover smaller solid angles but are over-represented in the image domain. Standard photometric losses therefore overweight polar regions, making optimization sensitive to noise and distortion. To address this, we propose a Panoramic Loss that weights each pixel by the cosine of its spherical latitude. This weighting explicitly accounts for the true area each pixel represents on the unit sphere, ensuring that the loss is consistent with spherical geometry rather than the distorted image domain:
\begin{equation}
\mathcal{L}_{pano} =
\sum_{\mathbf{u}}
\cos\big(\theta(\mathbf{u})\big)
\left\lVert
I_{\text{render}}(\mathbf{u}) -
I_{\text{gt}}(\mathbf{u})
\right\rVert.
\end{equation}
where $\theta(\mathbf{u})$ denotes the latitude angle of pixel $\mathbf{u}$ on the unit sphere. By enforcing area-consistent error accumulation, this sphere-aware loss significantly improves tracking stability.

\textbf{Keyframe Management.}  
While the front-end processes all incoming frames to maintain trajectory continuity, the back-end cannot optimize over every frame due to computational constraints. Therefore, the front-end is responsible for selecting keyframes and only forwards these to the back-end for optimization.
Following the keyframe management strategy of MonoGS\cite{Matsuki:Murai:etal:CVPR2024}, we select keyframes based on Gaussian point co-visibility and camera pose displacement. After keyframe selection, the rendered opacity map is sent to the backend to guide point cloud sampling.

\subsection{SLAM Back-end}

In the back-end mapping stage, we jointly optimize camera poses and Gaussian parameters by minimizing the photometric error. The optimization spans all frames in the keyframe window, along with two randomly sampled non-keyframe views, to alleviate global forgetting and enforce cross-view consistency. We reuse the panoramic loss ${L}_{pano}$ for back-end supervision. Furthermore, we propose a Depth-Guided Gaussian Initialization Strategy (DGIS), which consists of map initialization and Gaussian insertion, to provide better geometric priors and improve structural accuracy.

\begin{table*}[t]
\vspace{8pt}
\centering
\caption{Trajectory Accuracy Comparison on PALVIO Dataset\cite{wang2022lf} (ATE RMSE [m]).}
\vspace{-4pt}
\setlength{\tabcolsep}{5pt}
\label{tab:ate_PAL}
\begin{tabular}{llcccccccccc}
\toprule
\rowcolor{gray!10}
\textbf{Category} & \textbf{Method} 
& \textbf{ID01} & \textbf{ID02} & \textbf{ID03} & \textbf{ID04} & \textbf{ID05} & \textbf{ID06} & \textbf{ID07} & \textbf{ID08} & \textbf{ID09} & \textbf{ID10} \\
\midrule
\multirow{4}{*}{\textbf{Sparse Slam}}
& VINS-Mono$^{*}$\cite{qin2017vins}   
& / & 0.368 & 0.338 & 0.331 & 0.287 & / & 0.519 & / & / & / \\

& ORB-SLAM3$^{*}$\cite{ORBSLAM3_TRO} 
& 0.554 & 0.102 & 0.114 & 0.089 & 0.095 & 0.094 & 0.378 & 1.956 & 2.055 & 1.868 \\

& LF-VISLAM$^{*}$\cite{LF-VISLAM}
& / & 0.148 & 0.180 & 0.198 & 0.178 & 0.203 & 0.230 & 0.199 & 0.219 & / \\

& P2U-SLAM\cite{zhang2026p2u}
& \underline{0.108} & 0.073 & \underline{0.081} & \underline{0.086} & \underline{0.082}& 0.091& \underline{0.110}& \underline{0.109}& \underline{0.102}& \underline{0.109} \\

\midrule
\multirow{3}{*}{\textbf{GS-based Slam}}
& Photo-SLAM\cite{hhuang2024photoslam}
& / & \underline{0.068} & 0.094 & / & 0.129 & \underline{0.087} & 1.468 & 0.854 & 0.330 & 0.641 \\

& MonoGS\cite{Matsuki:Murai:etal:CVPR2024}
& 2.664 & 1.490 & 2.031 & 1.417 & 2.044 & 1.501 & 2.376 & 1.096 & 2.070 & 2.225 \\

& \textbf{Ours}  
& \textbf{0.078} & \textbf{0.055} & \textbf{0.071} & \textbf{0.074} & \textbf{0.068} & \textbf{0.076} & \textbf{0.091} & \textbf{0.091} & \textbf{0.102} & \textbf{0.102} \\
\bottomrule
\end{tabular}

\vspace{2pt}
\footnotesize{
Best results are in \textbf{bold}, second-best in \underline{underlined}.
“/” indicates failure.
$^{*}$ indicates methods with IMU. }
\end{table*}

\textbf{Map Initialization.}
Upon receiving the first frame in the back-end, we employ a pre-trained panoramic depth estimation network\cite{BiFuse20} to predict an initial depth map, which enables the construction of a reliable Gaussian map.
To initialize the map, Gaussian primitives are generated by downsampling the RGB-D observations. Specifically, we adopt an equal-area uniform sampling strategy on the unit sphere. This design alleviates Gaussian redundancy near the poles caused by equirectangular projection and ensures a more balanced spatial distribution across the panoramic field of view.

\textbf{Gaussian Insertion.}
For each incoming keyframe, additional Gaussian primitives are inserted to progressively model newly observed regions. To reduce redundancy, the insertion density is adaptively controlled based on the opacity map, built upon the unit-sphere sampling scheme. Regions with opacity values below a predefined threshold are considered under-reconstructed and are therefore assigned a higher insertion density. 
The depth of newly inserted Gaussians is initialized from the rendered depth map of the current frame. If no valid depth is available at the corresponding pixel, the depth is assigned from its nearest valid neighbor, followed by a small random perturbation. 
This strategy preserves geometric consistency with the current reconstruction while introducing sufficient diversity to enhance optimization robustness and reduce the risk of convergence to poor local minima.

\begin{table}[t]

\centering
\caption{Trajectory Accuracy Comparison on SynPano Dataset\cite{synpano2026} (ATE RMSE [m]).}
\vspace{-4pt}
\setlength{\tabcolsep}{5pt}
\label{tab:ate_syn}
\begin{tabular}{lccccc}
\toprule
\rowcolor{gray!10}
\textbf{Method} & \textbf{room1} & \textbf{room2} & \textbf{room3} & \textbf{room4} & \textbf{room5} \\
\midrule
ORB-SLAM3\cite{ORBSLAM3_TRO}  & 0.015 & 1.827 & \underline{0.023} & / & 1.366 \\

P2U-SLAM\cite{zhang2026p2u}  & 0.020 & \underline{0.135} & 0.066 & \underline{0.011} & \underline{0.020} \\
\midrule
Photo-SLAM\cite{hhuang2024photoslam}      & \underline{0.015} & 2.164 & 0.049 & / &/  \\

MonoGS\cite{Matsuki:Murai:etal:CVPR2024}           & 0.237 & 1.190 & 0.354 & 0.317 & 1.058 \\
\textbf{Ours}    & \textbf{0.005} & \textbf{0.018} & \textbf{0.003} & \textbf{0.003} & \textbf{0.010} \\
\bottomrule
\end{tabular}

\vspace{2pt}
\footnotesize{
VINS-Mono and LF-VISLAM require IMU data, which SynPano does not provide, so their results are unavailable.
}

\end{table}

\begin{figure*}[thpb]
  \centering
  \includegraphics[width=\linewidth]{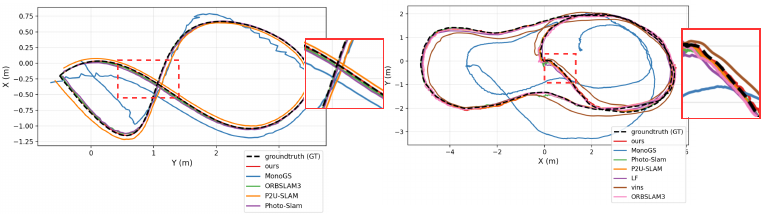}
  \caption{\textbf{Trajectory Comparison Against Ground Truth.}
    \textbf{Left:} room3 from the SynPano dataset. \textbf{Right:} ID02 from the PALVIO dataset.
    Our method achieves substantially better alignment with the ground truth than all baselines. In contrast, MonoGS (our primary baseline) suffers from unstable front-end tracking and inconsistent trajectory scale over time.}
  \label{fig:traj}
\end{figure*}
\section{EVALUATION}
\subsection{Experimental Setting}

\textbf{Baselines.}
We benchmark our method against classical geometric systems, including ORB-SLAM3\cite{ORBSLAM3_TRO} and VINS-Mono\cite{qin2017vins}, and their wide-FoV extensions P2U-SLAM\cite{zhang2026p2u} and LF-VISLAM\cite{LF-VISLAM}. 
We also compare with Gaussian-based SLAM frameworks, specifically MonoGS\cite{Matsuki:Murai:etal:CVPR2024} and Photo-SLAM\cite{hhuang2024photoslam}.
Our method builds upon the Gaussian representation as applied in MonoGS, while extending it to panoramic modeling and improving tracking robustness. Therefore, MonoGS serves as the primary baseline for evaluating the effectiveness of our proposed extensions.

\textbf{Datasets.}
We evaluate our method on the SynPano\cite{synpano2026} and PALVIO datasets\cite{wang2022lf}.
The SynPano dataset is synthetically generated using Blender and consists of indoor room scenes in the form of $360^\circ \times 180^\circ$ equirectangular panoramas. It provides a controlled environment to systematically investigate the impact of sensing geometry on optimization stability.
The PALVIO dataset features $360^\circ \times [40^\circ, 120^\circ]$ panoramic annular lens (PAL) sequences. It is captured by a real aerial platform and poses significant challenges for tracking due to aggressive 6-DOF maneuvers and rapid viewpoint changes.

\textbf{Data Preprocessing.}
To accommodate the diverse camera models of the baselines, we implement a flexible preprocessing interface tailored to each dataset's native format. 
For the SynPano dataset (native equirectangular), we project the panoramas into virtual pinhole views for pinhole-based systems and into Panoramic Annular Lens (PAL) images for those using the Scaramuzza (Taylor) model. 
Conversely, for the PALVIO dataset (native PAL), we map the raw frames into equirectangular images for our PanoGS-SLAM and into pinhole views for the other baselines. 
Crucially, all virtual pinhole views ($120^\circ \times 60^\circ$ FoV) are extracted from the equatorial region to minimize resampling artifacts, ensuring a rigorous and fair evaluation across different sensing geometries.
ORB-SLAM3, MonoGS, and Photo-SLAM use pinhole views as input, while other baselines use panoramic inputs.

\textbf{Metrics.}
We evaluate tracking accuracy using the root mean squared error (RMSE) of the absolute trajectory error (ATE). To account for scale ambiguity, the estimated trajectory is aligned with the ground truth before evaluation.
For reconstruction quality, we report photometric metrics including PSNR, SSIM, and LPIPS to evaluate the absolute rendering quality and photometric consistency of the reconstructed Gaussian maps.

\textbf{Computational Platform.} 
All experiments are conducted on a cloud server equipped with a single NVIDIA GeForce RTX 4090 GPU and 16 vCPUs based on an Intel(R) Xeon(R) Gold 6430 processor.
\begin{figure*}[thpb]
  \centering
  \includegraphics[width=\linewidth]{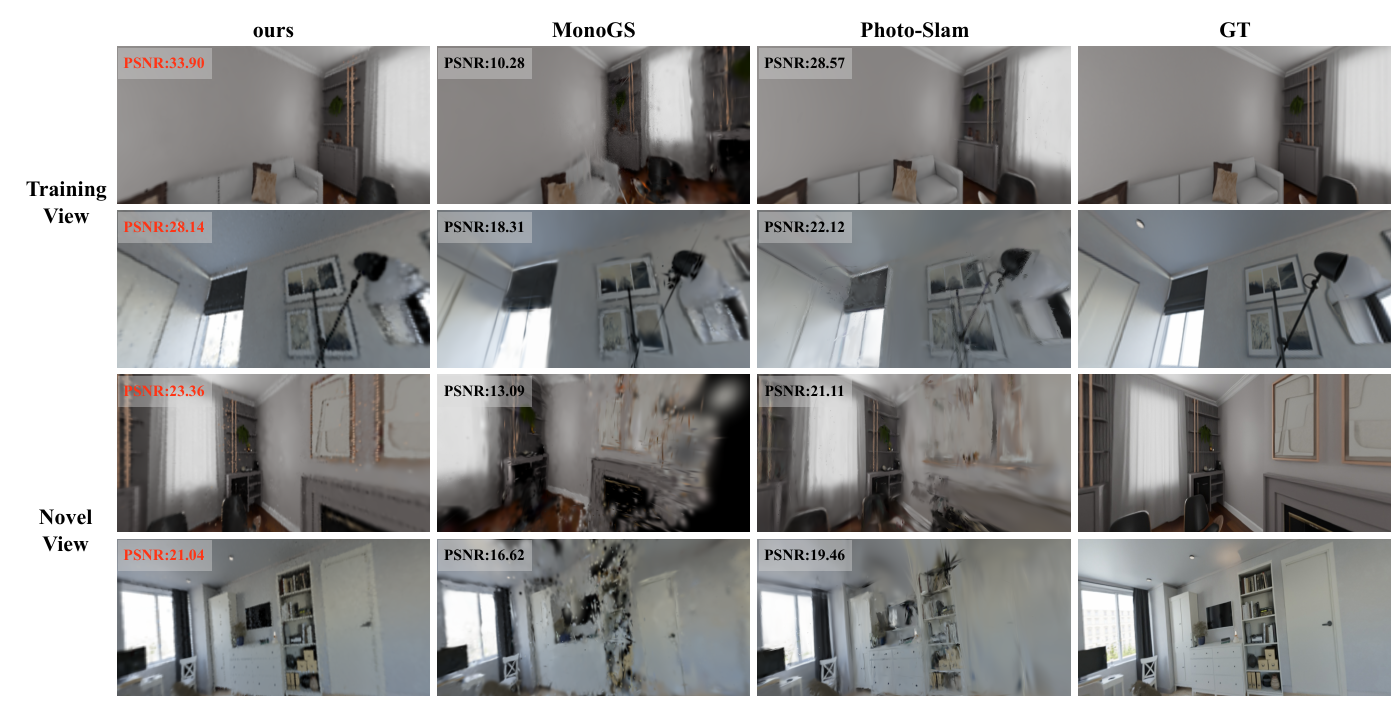}
    \caption{\textbf{Comparison of Rendering Quality in Training and Novel Views on SynPano.} MonoGS shows misaligned reconstructions due to trajectory errors, while Photo-SLAM exhibits artifacts and overfitting in novel views. The black regions in the MonoGS renderings do not correspond to unknown areas, as evidenced by the Photo-SLAM renderings.}
  \label{render}
\end{figure*}
\begin{figure}[thpb]
  \centering
  \includegraphics[width=\linewidth]{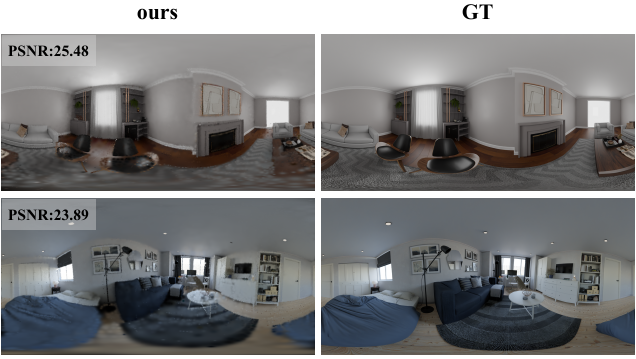}
    \caption{\textbf{Panoramic Rendering Examples.} 
    Our method produces high-quality panoramic renderings. }
  \label{render_pano}
\end{figure}

\subsection{Quantitative Evaluation}
\textbf{Tracking Accuracy.}
As summarized in Tables~\ref{tab:ate_PAL} and~\ref{tab:ate_syn}, PanoGS-SLAM consistently achieves the lowest ATE RMSE across all evaluated sequences. 
While traditional geometric baselines such as LF-VISLAM and P2U-SLAM leverage wide-FoV inputs, their reliance on discrete feature matching can lead to instability under aggressive 6-DOF maneuvers.
In contrast, PanoGS-SLAM maintains stable tracking by leveraging continuous photometric gradients from our differentiable rendering pipeline, which provides more resilient pose constraints during rapid viewpoint changes on the PALVIO benchmark (Tables~\ref{tab:ate_PAL}).
Compared to Gaussian-based dense SLAM systems, our method demonstrates superior trajectory alignment (Fig.~\ref{fig:traj}) by resolving the weakened rotational observability and poorly conditioned optimization inherent to narrow-FoV sensing. 
This advantage is most pronounced in SynPano’s room4 and room5 sequences  (Tables~\ref{tab:ate_syn}), where $360^\circ \times 180^\circ$ coverage ensures reliable convergence in texture-less regions where pinhole-based systems typically diverge.

\textbf{Rendering Quality.}
Table~\ref{tab:psnr} reports rendering quality compared with GS-based SLAM systems, where our method consistently outperforms all baselines. 
Fig.~\ref{render} shows visual comparisons where novel views are rendered from random poses outside the sequence. In these cases, MonoGS suffers from map distortions due to trajectory drift, while Photo-SLAM produces artifacts due to elongated Gaussians and overfitting. Notably, despite being trained with a panoramic model, our method still outperforms the baselines even on individual pinhole views corresponding to their input observations. 
Fig.~\ref{render_pano} presents panoramic renderings, showing geometrically consistent and visually coherent reconstructions under full panoramic projection, though the larger FoV naturally yields lower PSNR than pinhole renderings.

\begin{table}[!t]

\centering
\caption{Average Rendering Quality Comparison.}
\vspace{-4pt}
\setlength{\tabcolsep}{4pt}
\label{tab:psnr}
\begin{tabular}{llccc}
\toprule
\rowcolor{gray!10}
\textbf{Dataset} & \textbf{Method} & \textbf{PSNR(dB) $\uparrow$} & \textbf{SSIM $\uparrow$} & \textbf{LPIPS $\downarrow$} \\
\midrule
\multirow{3}{*}{\textbf{PALVIO}\cite{wang2022lf}}
& Photo-SLAM\cite{hhuang2024photoslam} & 20.10 & 0.83 & 0.40 \\
& MonoGS\cite{Matsuki:Murai:etal:CVPR2024}     & 18.17 & 0.79 & 0.47 \\
& \textbf{Ours} & \textbf{23.48} & \textbf{0.88} & \textbf{0.30} \\
\midrule
\multirow{3}{*}{\textbf{SynPano}\cite{synpano2026}}
& Photo-SLAM\cite{hhuang2024photoslam} & 24.46 & 0.82 & 0.30 \\
& MonoGS\cite{Matsuki:Murai:etal:CVPR2024}     & 21.47 & 0.81 & 0.31 \\
& \textbf{Ours} & \textbf{30.58} & \textbf{0.91} & \textbf{0.22} \\
\bottomrule

\end{tabular}

\vspace{2pt}
\footnotesize{Direct comparison between panoramic and pinhole renderings is non-trivial. To ensure a fair comparison with pinhole-based baselines, all methods are evaluated using pinhole rendering; three $120^\circ$ FoV views (forward/left/right) are rendered and averaged at each pose. For SynPano \textit{room1}, only the forward view is used since the FoV of pinhole baselines does not cover the entire scene.}
\vspace{-4pt}
\end{table}

\textbf{Ablative Analysis.}
Table~\ref{tab:ablation} presents an ablation study assessing the effectiveness of our proposed unit-sphere panoramic loss $\mathcal{L}_{pano}$ and depth-guided Gaussian initialization strategy (DGIS). The unit-sphere loss $\mathcal{L}_{pano}$ aligns geometrically with panoramic projection, effectively mitigating noise amplification near the poles, and thus substantially reduces trajectory errors across all sequences.
For comparison, $\mathcal{L}_{GS}$ employs a Gaussian weighting scheme that decreases penalties from the equator toward the poles, serving as a baseline to confirm the geometric superiority of our cosine-based $\mathcal{L}_{pano}$.
Our DGIS yields a more reliable initial Gaussian map and enhances the quality of newly inserted Gaussians during incremental mapping, further lowering trajectory errors in most sequences.

\begin{table}
\vspace{8pt}
\centering
\caption{Ablation Study on SynPano Dataset\cite{synpano2026}.}
\vspace{-4pt}
\label{tab:ablation}
\begin{tabular}{ccccc}
\toprule
\rowcolor{gray!10}
\textbf{Method} &
\textbf{ATE(cm) $\downarrow$} &
\textbf{PSNR(dB) $\uparrow$} &
\textbf{SSIM $\uparrow$} &
\textbf{LPIPS $\downarrow$}
\\ \midrule
w/o ${\mathcal{L}_{pano}}$
 & 5.19 & 27.23 & 0.86 &  0.30  \\
w ${\mathcal{L}_{GS}}$
& 0.84 & 29.88    & 0.91 &  0.22  \\
w/o DGIS    
 & 1.27 & 28.68    & 0.89 &  0.24   \\
\textbf{Ours}   
 & \textbf{0.78}  & \textbf{30.58} & \textbf{0.91} & \textbf{0.22} \\
\bottomrule
\end{tabular}

\vspace{2pt}
\footnotesize{DGIS denote Depth-Guided Gaussian Initialization Strategy. ${\mathcal{L}_{GS}}$ is a Gaussian-weighted loss whose weights decrease progressively from the equator to the poles. }
\vspace{-4pt}
\end{table}
\begin{figure}[t]
  \centering
  \includegraphics[width=\linewidth]{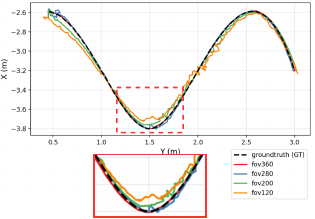}
  \caption{\textbf{Trajectory Comparison under Different FoVs.} Larger FoV lead to smoother trajectories and closer alignment with the ground truth.}
  \label{fig:fov_traj}
\vspace{-4pt}
\end{figure}

\begin{table}[t]
\vspace{8pt}
\centering
\caption{FOV Study on SynPano Dataset\cite{synpano2026} (ATE RMSE [cm]).}
\vspace{-4pt}
\label{tab:fov}
\setlength{\tabcolsep}{5pt}
\begin{tabular}{c c ccccc}
\toprule
\rowcolor{gray!10}
\textbf{Method} & \textbf{FOV} &
\textbf{room1} & \textbf{room2} & \textbf{room3} & \textbf{room4} & \textbf{room5} \\
\midrule
MonoGS\cite{Matsuki:Murai:etal:CVPR2024} 
& $120^\circ$ 
& 23.71 & 118.98 & 35.45 & 31.38 & 105.80 \\
\midrule
\multirow{4}{*}{Ours} 
& $120^\circ$ & 5.26 & 59.64 & 102.28 & 25.82 & 19.96 \\
& $200^\circ$ & 2.89 & 11.29 & 0.86 & 6.34 & 6.27 \\
& $280^\circ$ & 1.36 & 2.48 & 0.34 & 0.81 & 4.73 \\
& $360^\circ$ & \textbf{0.45} & \textbf{1.79} & \textbf{0.26} & \textbf{0.21} & \textbf{1.08} \\
\bottomrule
\end{tabular}
\vspace{-4pt}
\end{table}
\textbf{Effect of Field-of-View.}
Table~\ref{tab:fov} presents a controlled FoV study on the SynPano dataset. Fig.~\ref{fig:fov_traj} illustrates detailed trajectory comparisons against the ground truth under different FoV settings. Results indicate a clear monotonic trend: trajectory accuracy improves markedly as the visible angular range increases across all sequences. Notably, challenging sequences (e.g., room2 and room3) degrade severely at a narrow FoV (120°), but they experience a critical performance turning point when the FoV exceeds 200°. 
This phenomenon suggests that wide-field sensing fundamentally alters the conditioning of the underlying pose optimization problem. Under limited FoV, photometric gradients are concentrated within a narrow angular region, leading to insufficient rotational observability and unstable optimization. As the FoV expands, gradient contributions are distributed across a broader spherical domain, significantly improving numerical stability and convergence behavior. Furthermore, under the same FoV setting, our method consistently outperforms MonoGS across all four scenes.

\begin{table}

\centering
\caption{Convergence Behavior under Different Iteration Counts on PALVIO Dataset\cite{wang2022lf}.}
\vspace{-4pt}
\label{tab:fps}
\begin{tabular}{lccccc}
\toprule
\multirow{2}{*}[-0.6ex]{\textbf{Method}} 
& \multirow{2}{*}[-0.6ex]{\textbf{Metric}} 
& \multicolumn{4}{c}{\textbf{Front-end / Back-end Iterations}} \\
\cmidrule(lr){3-6}
& & 15/30 & 30/50 & 50/80 & 100/100 \\
\midrule
\multirow{2}{*}{MonoGS\cite{Matsuki:Murai:etal:CVPR2024}}        
& ATE(m) $\downarrow$  & 2.314   & 1.872   & 1.566  & 1.490    \\
& FPS $\uparrow$   & 5.07      & 3.22      & 2.13      & 1.19       \\ \midrule
\multirow{2}{*}{\textbf{Ours}}   
& ATE(m) $\downarrow$   & 0.056 & 0.054 & 0.052 & 0.053 \\
& FPS $\uparrow$ & 7.04 & 4.47 &  3.16 & 1.46 
 \\ \bottomrule
\end{tabular}
\vspace{-4pt}
\end{table}
\begin{figure}[!h]
  \centering
  \includegraphics{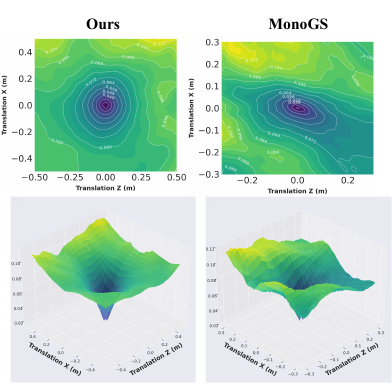}
\caption{\textbf{Convergence Basin On SynPano Dateset.}
\textbf{Top:} Loss contour map on the X–Z plane.
\textbf{Bottom:} 3D visualization of the X–Z–Loss surface. Our PanoGS-SLAM shows faster and more robust convergence with a wider attraction basin and consistent gradients.
}
  \label{basin}
\vspace{-4pt}
\end{figure} 

\textbf{Pose Optimization Convergence.}
In Table~\ref{tab:fps}, we observe that PanoGS-SLAM converges significantly faster during camera pose optimization. Specifically, PanoGS-SLAM achieves pose convergence within only 15 optimization iterations, reaching 7 FPS, whereas MonoGS exhibits convergence trends only after approximately 100 iterations.
To further analyze the reasons for more stable and faster optimization, we conduct a convergence basin study by perturbing the camera pose around the ground-truth pose. Perturbations are scaled relative to the median scene depth (with a factor of 0.5) to maintain consistency across diverse environments. As illustrated in Fig~\ref{basin}, PanoGS-SLAM manifests a significantly broader and more well-conditioned basin of attraction within the optimization landscape, ensuring enhanced convexity and robust convergence compared to the narrow-FoV MonoGS. 
This improvement stems from the synergy between our native spherical representation and the differentiable rendering pipeline.
Unlike pinhole models, which clip gradients at image boundaries, our panoramic framework maintains a continuous gradient field over the entire viewing sphere. This ensures strong gradient consistency in all directions, as gradients are not truncated in space, making optimization easier while enhancing pose observability and enabling faster, more stable convergence.
                        
\section{Conclusion}

We presented PanoGS-SLAM, the first panoramic SLAM framework that jointly performs camera tracking and 3D Gaussian map optimization directly within the spherical domain. To address the inherent distortions of equirectangular projections and maintain strict geometric consistency, we introduced a sphere-consistent photometric objective alongside a depth-guided Gaussian initialization strategy. Furthermore, our work reveals a fundamental connection between sensing geometry and numerical robustness in differentiable SLAM. Future research will explore the integration of global optimization and loop closure for large-scale, long-term deployments.

\bibliographystyle{IEEEtran}
\bibliography{refs}

\end{document}